\documentclass[conference]{IEEEtran}
\IEEEoverridecommandlockouts
\usepackage{cite}

\usepackage{amsmath,amssymb,amsfonts}
\usepackage{algorithm}
\usepackage{algorithmic}
\usepackage{graphicx}
\usepackage{textcomp}
\usepackage{xcolor}
\usepackage{makecell}
\usepackage{multicol}
\usepackage{multirow}
\usepackage{bm}
\usepackage{colortbl}
\usepackage{array}
\usepackage{pifont}
\usepackage{float}
\usepackage{xcolor}
\usepackage{lineno}
\usepackage{amssymb}
\usepackage[hyphens]{url}

\definecolor{sota}{RGB}{0, 0, 255}
\usepackage[colorlinks, citecolor=green]{hyperref}

\usepackage{natbib}
\setcitestyle{numbers,square}
\usepackage{booktabs}
\usepackage{subcaption}
\usepackage{enumitem}

\usepackage{tcolorbox}
\usepackage{natbib}
\def\BibTeX{{\rm B\kern-.05em{\sc i\kern-.025em b}\kern-.08em
    T\kern-.1667em\lower.7ex\hbox{E}\kern-.125emX}}
\begin{document}

\title{CLIP Brings Better Features to Visual Aesthetics Learners}

\author{
 Liwu Xu\textsuperscript{1*},
 Jinjin Xu\textsuperscript{1*\dag},
 Yuzhe Yang\textsuperscript{1},
 Xilu Wang\textsuperscript{2},
 Yi-Jie Huang\textsuperscript{1},
 Yaqian Li\textsuperscript{1\dag} \\

\textsuperscript{1} OPPO AI Center \textsuperscript{2} University of Surrey

}

\maketitle

\def\thefootnote{}\footnotetext{\hrule width 0.62\columnwidth height 0.5pt \vspace{2pt} 
  \hangindent=\parindent \textsuperscript{*}Equal contribution. ~\textsuperscript{\dag}Corresponding authors. \\
  \hangindent=\parindent \{jinxu95, liyaqian\}@oppo.com, ICME 2025}

\begin{abstract}
Image Aesthetics Assessment (IAA) is a challenging task due to its subjective nature and expensive manual annotations. Recent large-scale vision-language models, such as Contrastive Language-Image Pre-training (CLIP), have shown their promising representation capability for various downstream tasks. However, the application of CLIP to resource-constrained and low-data IAA tasks remains limited. While few attempts to leverage CLIP in IAA have mainly focused on carefully designed prompts, we extend beyond this by allowing models from different domains and with different model sizes to acquire knowledge from CLIP. To achieve this, we propose a unified and flexible two-phase \textbf{C}LIP-based \textbf{S}emi-supervised \textbf{K}nowledge \textbf{D}istillation (CSKD) paradigm, aiming to learn a lightweight IAA model while leveraging CLIP's strong generalization capability. Specifically, CSKD employs a feature alignment strategy to facilitate the distillation of heterogeneous CLIP teacher and IAA student models, effectively transferring valuable features from pre-trained visual representations to two lightweight IAA models, respectively. To efficiently adapt to downstream IAA tasks in a low-data regime, the two strong visual aesthetics learners then conduct distillation with unlabeled examples for refining and transferring the task-specific knowledge collaboratively. Extensive experiments demonstrate that the proposed CSKD achieves state-of-the-art performance on multiple widely used IAA benchmarks. Furthermore, analysis of attention distance and entropy before and after feature alignment shows the effective transfer of CLIP's feature representation to IAA models, which not only provides valuable guidance for the model initialization of IAA but also enhances the aesthetic feature representation of IAA models. Code will be made publicly available.
\end{abstract}


\subsection{Introduction}
\label{sec:intro}
Image aesthetics assessment (IAA) aims at evaluating the aesthetics perception of an image automatically and has been broadly applied in various areas, such as image enhancement \cite{talebi2018learned}, personalized photo album management \cite{mai2016composition} and etc. It is widely acknowledged that capturing aesthetic features for IAA is challenging due to the highly subjective and ambiguous nature of the definition of aesthetics \cite{hosu2019effective,ke2021musiq}. Early approaches for IAA mainly represent aesthetic patterns on hand-crafted features \cite{ke2006design}, which encounter difficulties in capturing complicated human visual perception. With the advances of deep learning (DL), DL-based approaches for extracting aesthetic features for IAA have emerged and shown promising performance in recent years, such as multi-patch aggregation \cite{9777255}, composition-aware IAA \cite{she2021hierarchical} and multi-scale feature fusion \cite{ke2021musiq}. Among them, considerable efforts have been devoted to leveraging large-scale pre-trained models, such as the ImageNet classification task, to learn rich transferable knowledge \cite{ke2021musiq,hou2022distilling}. 


More recently, large-scale multi-modal pre-trained models, such as GPTs \cite{brown2020language}, CoCa \cite{yu2022coca} and CLIP \cite{radford2021learning}, have shown remarkable zero-shot transfer and general representation ability. In light of this, several attempts have been done to explore the use of vision-language pre-training methods in IAA \cite{ wang2023exploring,ke2023vila,zhang2023blind}, successfully demonstrating their effectiveness for extracting useful aesthetic features. For example, Ke \emph{et al.} \cite{ke2023vila} proposed to pre-train a vision-language model by leveraging image-comment pairs, followed by fine-tuning a lightweight rank-based adapter while freezing the pre-trained weights. Their studies revealed that CLIP could provide a wide range of useful features for IAA, such as lighting, composition, and properties of beauty ideals, especially compared with pre-training on an ImageNet classification task. Alternatively, Wang \emph{et al.} \cite{wang2023exploring} utilized CLIP for visual perception using prompts to describe image contents. Instead of using prompt tuning,  a pre-trained CLIP model for feature embeddings is fine-tuned using a multi-task learning approach to enhance image quality assessment \cite{zhang2023blind}. 

While the existing work has shed light on the representation ability of CLIP for IAA \cite{wang2023exploring,ke2023vila}, there remain several challenges in this line of research:  1) Vision-language models may under-represent aesthetic-related information, as they are typically pre-trained on image-text pairs that focus on aligning feature spaces to capture rich semantic information across modalities. Consequently, it is necessary to fine-tune the task-agnostic pre-trained model into a task-specific one to acquire domain-specific knowledge \cite{chen2020big}. 2) A direct adaptation of CLIP by fine-tuning on data-scarce IAA tasks may fail to capture the unique characteristics of image aesthetics, hindering the sub-optimal transferability \cite{li2024graphadapter}. Notably, applying an MLP on the top of the CLIP image encoder has successfully leveraged the multimodal foundation model, such as WiSE-FT \cite{wortsman2022robust} and VILA \cite{ke2023vila}. However, this approach is sensitive to the hyperparameters. As shown in Table \ref{tab:para}, we fine-tuned CLIP with varying learning rates, i.e., $5\times 10^{-5}$, $1\times 10^{-5}$, $5\times 10^{-6}$, while keeping other hyperparameters fixed on PARA benchmark \cite{yang2022personalized}. The corresponding SRCC and PLCC metrics varied significantly, indicating that this approach often suffers from training instability and inefficiency issues. 3) Deploying DL-based IAA models on mobile devices remains problematic due to the high space and time complexity, and most lightweight designs often compromise performance significantly.
\begin{figure*}[thb]
\centering
\includegraphics[width=0.9\linewidth]{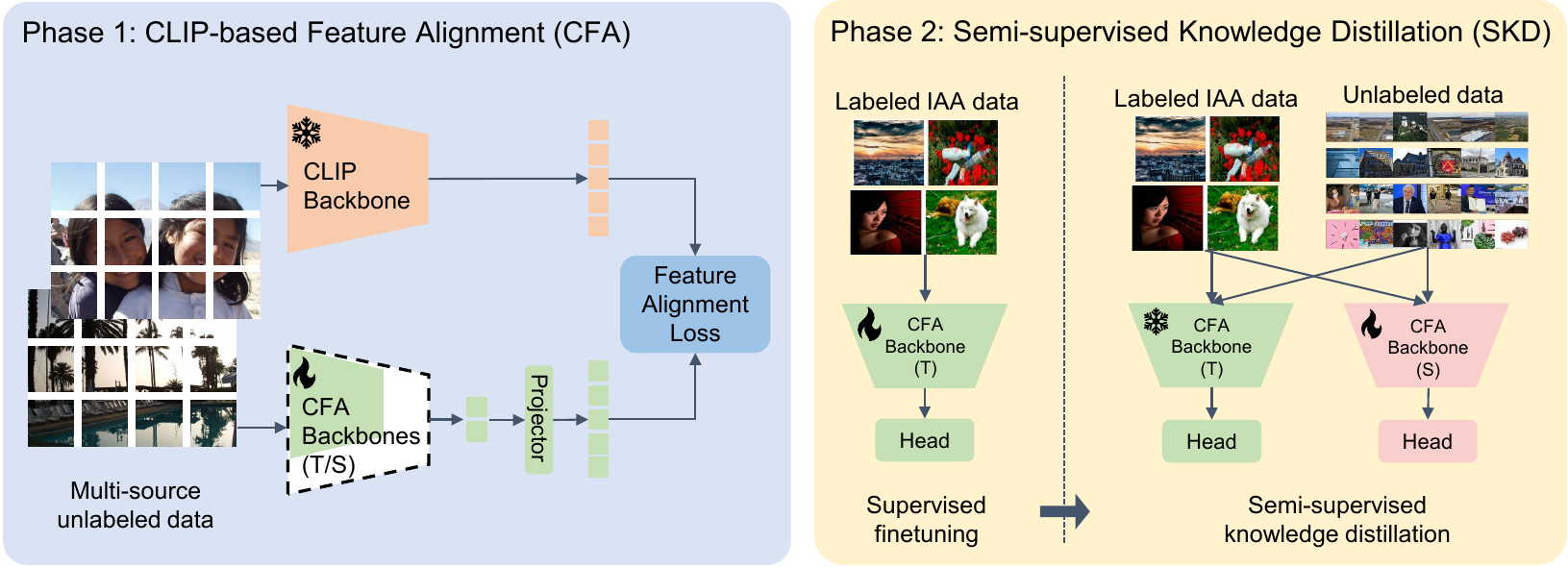}
\caption{The overall workflow of CSKD includes two main phases: CFA and SKD. In CFA phase, transferable knowledge is distilled from the CLIP teacher to two ImageNet pre-trained backbones using a feature alignment strategy on a multi-source unlabeled dataset, resulting two CFA models. In SKD phase, one of the CFA models is fine-tuned on the IAA dataset as a teacher model, then transfers knowledge to the other CFA model via a semi-supervised knowledge distillation strategy.}
\label{fig:pipeline}
\end{figure*}

In this work, we aim to leverage the representation ability of CLIP while maintaining a lightweight IAA model to overcome the above-mentioned challenges. A straightforward way to achieve this is to directly distill transferable knowledge from CLIP to a small on-device IAA model. Unfortunately, such a lightweight student model exhibits only marginal performance improvements when trained on limited datasets, as shown in Table \ref{tab:ava}. This phenomenon is commonly observed in computer vision tasks due to the low information density of visual inputs. Based on these observations, we hypothesize that 1) CLIP  can provide better aesthetic features for the IAA downstream task, and 2) a small on-device IAA model can be obtained via knowledge distillation, leveraging the learned representations from the pre-training phase, but 3) such a model can be further improved via fine-tuning with task-specific unlabeled data. According to these hypothesis, we propose a novel \textbf{C}LIP-based two-phase \textbf{S}emi-supervised \textbf{K}nowledge \textbf{D}istillation method for IAA, termed \textbf{CSKD}, to conduct knowledge distillation along with semi-supervised learning.

As shown in Fig.\ref{fig:pipeline}, CSKD firstly utilizes a large amount of unlabeled data ($\approx$3 million) to align the features of the visual encoders and CLIP image encoder with no constraints on model structure or size. Then, the well-trained encoder (e.g., Swin-Base) is combined with a task-specific MLP head and fine-tuned on image aesthetics datasets in a supervised manner, which leads to a strong teacher for pseudo label generation. Finally, we train an efficient and generalizable student, MobileNetV2 (MV2) \cite{sandler2018mobilenetv2}, via semi-supervised knowledge distillation. Contributions can be summarized as below:

\begin{itemize}
    \item To leverage the powerful feature representation ability of CLIP,  we propose a unified and generalizable unsupervised \textbf{C}LIP-based \textbf{F}eature \textbf{A}lignment (CFA) method. This approach distills knowledge from CLIP to lightweight IAA models. The resulting model demonstrates superior zero-shot performance on IAA tasks, validating CFA's effectiveness as a pre-training scheme for visual aesthetics encoders.
    \item To further enhance the CFA model's performance on data-scarce IAA tasks, a \textbf{S}emi-supervised \textbf{K}nowledge \textbf{D}istillation (SKD) strategy is proposed, allowing large-scale unlabeled data to be used for learning the task-specfic knowledge. Our experiments demonstrate that both teacher and student models distilled from CLIP outperform their supervised counterparts with the same model structure.
    \item We evaluate our proposed pipeline against state-of-the-art (SOTA) IAA methods on various commonly used IAA benchmarks. The extensive experimental results demonstrate that our CSKD approach achieves superior performance on IAA tasks. An analysis of attention distance and entropy before and after feature alignment in Supplementary Material demonstrates the effective transfer of CLIP's feature representation capabilities.

\end{itemize}

\section{Related Work}

\subsection{Pre-training in IAA}
Pre-training has been a dominant paradigm to extract general features and has been proven beneficial for many computer vision tasks \cite{deng2009imagenet}. For example, ImageNet pre-training \cite{deng2009imagenet} is the most widely used general-purpose foundation model in the computer vision community. More recently, the success of natural language processing has fostered the development of large-scale vision-language pre-trained models. Among them, OpenAI's CLIP model \cite{radford2021learning} represents a crucial milestone due to its pioneering use of large-scale image-text data via contrastive learning, yielding better feature representation ability and impressive zero-shot performance. Inspired by CLIP's performance, the pre-trained CLIP visual encoder has been widely adopted as the feature extractor for various downstream tasks, such as MaskCLIP \cite{zhou2022extract} and Stable Diffusion \cite{rombach2022high}. 

Recently, there has been an emergence of adopting CLIP in the IAA domain, providing insights into the advantage of CLIP as an aesthetic feature extractor. However, most existing work relies on prompt engineering and requires text input during inference, necessitating computationally expensive fine-tuning on large datasets. Moreover, as stated in \cite{fang2023eva}, large CLIP models usually suffer from training instability and inefficiency issues. This limits its flexibility and application to on-device IAA. In this paper, we distill the transferable knowledge of CLIP to IAA models to leverage the representation ability of CLIP while maintaining shallower student models.

\subsection{Knowledge \& Feature Distillation}

Knowledge distillation (KD) \cite{hinton2015distilling} is a training paradigm that transfers knowledge of a larger teacher model to a smaller student model by using both ground truth labels and pseudo labels (hard or soft) generated by the teacher model. It is widely acknowledged that KD can benefit the student model in terms of the reduction of model size and the improvement of model generalization \cite{9340578}. Nevertheless, we observe that fine-tuning the student model obtained after KD suffers from the limited number of labeled data in IAA tasks, limiting the performance improvement (refer to Table E in Supplementary Material). Alternatively, Chen \emph{et al.} \cite{chen2020big} claimed that incorporating unlabeled data in knowledge distillation can significantly enhance learning efficiency. Therefore, given the scarcity of labeled samples and the abundance of unlabeled data in IAA, we propose to use a semi-supervised knowledge distillation method to further enhance the student model. 

Feature distillation (FD) is a subclass of KD, where knowledge is transferred by learning feature representations instead of outputs \cite{wang2021knowledge, yang2021survey}. FD is characterized by its flexibility that allows the teacher and student models to originate from different domains, possess varying output dimensions, and exhibit other dissimilarities. As a promising training paradigm, recent studies demonstrate that the strong vision representations learned by CLIP can be transferred to an untrained model via feature distillation \cite{fang2023eva}. Inspired by this, we adopt an off-the-shelf CLIP image encoder as a teacher to transfer rich knowledge to our given visual encoder. 

Notably, there are two major differences in our method. 1) Distillation target. We choose to use the CLS token, so that there is no model structure dependency and spatial alignment issues. 2) Initialization of distilled model. We start with ImageNet pre-trained parameters instead of training from scratch for faster convergence and better performance.

\section{CSKD: CLIP-based Semi-supervised KD}
As shown in Fig.\ref{fig:pipeline}, the overall pipeline of CSKD consists of two key phases: CLIP-based feature alignment (CFA) and semi-supervised knowledge distillation (SKD). CFA transfers feature representations from CLIP to backbone models, while SKD further refines the CFA backbones for downstream tasks. Phase I allows backbone models to learn from the CLIP backbone by aligning their features on a multi-source (e.g., ImageNet \& OpenImages detection subset) unlabeled dataset. In Phase I, two CFA backbones are trained separately to serve as the teacher and student models for the SKD phase. In phase II, the CFA teacher is firstly fine-tuned on task-specific data, followed by the training of the CFA student through semi-supervised knowledge distillation, leveraging both labeled and unlabeled data annotated by the
CFA teacher. The following sections detail the two core components of CSKD.

\subsection{Phase I: CLIP-based Feature Alignment (CFA)}
\label{sec:cfa}
This section details our use of the pre-trained CLIP image encoder to refine the feature representation of IAA task-specific backbones, as shown in the left panel of Fig. \ref{fig:pipeline}. As previously noted, CLIP's feature collapse issue has been mitigated through extensive training on image-text pairs. Our approach focuses on aligning feature representations rather than logit outputs, thus requiring only unlabeled data. 

Specifically, we construct a multi-source unlabeled dataset comprising approximately 3M images: 1.2M from ImageNet \cite{deng2009imagenet} and 1.9M from OpenImages \cite{kuznetsova2020open} detection subset. Given a pre-trained backbone (we adopt Swin-Base-In21k from timm \cite{rw2019timm}) as a student model, we can align the representation maps between the CLIP image encoder and the given backbone (followed by an MLP projector) via a cosine similarity based alignment loss,
\begin{equation}
    \mathcal{L}_{align}(\bm{x}_1, \bm{x}_2) = 1 - \frac{\bm{x}_1 \cdot \bm{x}_2}{max(||\bm{x}_1||_2 \cdot ||\bm{x}_2||_2, \epsilon)}, 
\end{equation}
where $\bm{x}_1$ represents the CLS token of CLIP,  $\bm{x}_2$ denotes the projected feature of the given backbone, and $\epsilon$ is a constant that is usually set to $1e-8$ to avoid zero dividing. Discussions about the feature representations before and after alignment are given in Section II.A in Supplementary Material.

Notably, we train different backbones (e.g., Swin-Base \& MV2 \cite{sandler2018mobilenetv2}) in this way for the following phase, referred to as CFA teacher and CFA student models, respectively.

\subsection{Phase II: Semi-supervised Knowledge Distillation (SKD)}
\label{sec:skd}
Fine-tuning is a widely adopted approach for adapting task-agnostic pre-trained models to specific tasks. Once the two CFA models are obtained, we perform the SKD illustrated in the right panel of Fig. \ref{fig:pipeline} to further improve the target model for IAA tasks. This process first fine-tunes one of the CFA models with an MLP head using AVA \cite{murray2012ava}, AADB \cite{kong2016photo} or PARA \cite{yang2022personalized} benchmarks to serve as a teacher model. Then, the well-trained teacher is used to supervise the student (another CFA backbone with an MLP head). To circumvent the low-data challenge encountered in IAA tasks, we fine-tune the student model on both labeled and unlabeled data by minimizing the squared earth-mover-distance (EMD) \cite{talebi2018nima} between predictions and human-annotated labels or pseudo labels,
\begin{equation}
    \mathcal{L}_{EMD}(\bm{p}, \bm{q}) = (\frac{1}{d}\sum_{i=1}^{d}|(CDF_{\bm{p}}(i)-CDF_{\bm{q}}(i)|^r)^{1/r}, 
\end{equation}
in which $r$ is a hyper-parameter ($r=2$ in our method), $\bm{p} \in R^d$ and $\bm{q} \in R^d$ represent prediction and target vectors respectively, and $CDF$ represents the cumulative distribution function. The objective function of the student model in the SKD is a combination of the knowledge distillation loss $\mathcal{L}_{kd}$ and the loss of ground-truth labeled data $\mathcal{L}_{s}$:
\begin{equation}
    \mathcal{L}_{stu} = \mathcal{L}_{s} + \beta * \mathcal{L}_{kd},
\end{equation}
where $\beta$ is a fixed balancing factor, and 
\begin{equation}
    \mathcal{L}_{s} = \frac{1}{B_s}\sum_{j=1}^{B_s}\mathcal{L}_{EMD}(\bm{p}^{l}, \bm{q}^{l}),
\end{equation}
with $B_{s}$ the size of labeled data per batch, and
\begin{equation}
    \mathcal{L}_{kd} = \frac{1}{B_s +\mu B_s}\sum_{k=1}^{B_s +\mu B_s}\mathcal{L}_{EMD}(\bm{p}^{u}, \bm{\hat{q}}^{u}),
\end{equation}
where $\mu$ indicates the proportion of 
unlabeled data, and $\bm{\hat{q}}^{u}$ is the pseudo target vector.

\section{Experiments}
\subsection{Implementation Details}
{\noindent\bf Datasets and Evaluation Protocol}
Following previous work, three commonly used IAA datasets, namely AVA \cite{murray2012ava}, AADB \cite{kong2016photo} and PARA \cite{yang2022personalized}, are adopted. AVA, AADB, and PARA contain 242,538, 10,000, and 31,220 samples labeled with mean opinion scores, respectively. From these datasets, 237,623, 8,345, and 28,220 samples are randomly selected for training, respectively.

{\noindent\bf Architectures and Training Configurations}
All experiments are conducted using 4 NVIDIA Tesla V100 GPUs and PyTorch framework. The pre-processing starts with scaling the images to a size of $256\times256$, followed by a random crop of size $224\times224$. Finally, a random horizontal flip is performed with a probability of 0.5. The default image encoder of CLIP used in the CFA phase is ViT-L-14. To conduct the SKD phase, the useful generalizable knowledge of CLIP is distilled to two IAA backbones: one is initialized with ImageNet-21k (Swin-Base) from timm \cite{rw2019timm} and serves as the teacher model in the SKD, and the other is initialized with ImageNet pre-trained weights (either MV2 \cite{sandler2018mobilenetv2} or ViT-Tiny \cite{touvron2021training}) and serves as the student model. The initial learning rate (lr) of Swin-Base, ViT-Tiny and MV2 is fixed to $1\times 10^{-4}$. The lr decays by a factor of 0.1 at epoch 5 for MV2 and ViT-Tiny, and at epoch 2 for Swin-Base, with a total of 16 epochs. We use Adam as the optimizer, and set both $\mu$ and $\beta$ in the SKD stage to 15. 

\begin{table*}[ht]
\centering
\caption{Results on the AVA dataset. Note that NIMA* represents our replicated supervised version using ImageNet (MV2, ViT-Tiny) and ImageNet-21k (Swin-Base) pre-trained backbones, and "full" means full resolution image, up to 800 × 800 pixels. Results in blue font color indicate the SOTA performance under this metric.}
\label{tab:ava}
\resizebox{0.89\linewidth}{0.22\linewidth}{
\begin{tabular}{l|cc|cc|ccc}
\toprule
Learning scheme &Methods & Year & \# Backbone & \# Input size  & MSE $\downarrow$ & SRCC $\uparrow$  & PLCC $\uparrow$ \\
\midrule
\multirow{13}{*}{\shortstack{Supervised}}
&MNA-CNN \cite{mai2016composition} & 2016 & VGG + Places205-GoogLeNet &   224      & / & /     & / \\
&Kong et al. \cite{kong2016photo}  & 2016 & AlexNet & 227                             & /     & 0.558     & / \\
&AMP \cite{murray2017deep}         & 2017 &  ResNet101 & full                         & 0.279 &    0.709       & / \\
&\shortstack{NIMA \cite{talebi2018nima}}       & \shortstack{2018}  &VGG16& 224                             & /     & 0.592 &0.610\\
& \shortstack{NIMA \cite{talebi2018nima}}        &\shortstack{2018}   & Inception-v2 & 224                       & /     & 0.612& 0.636\\
&MPADA \cite{sheng2018attention}   & 2018  &ResNet18  &224 [$\geq$ 32 crops]          & /& / & / \\
&Zeng et al \cite{zeng2019unified} & 2019  & ResNet101 & 384                          &0.275 & 0.719& 0.720\\
&Hosu et al. \cite{hosu2019effective} & 2019 & Inception-v3& (full*0.875) [20 crops] & / & {0.756} & 0.757 \\
&AFDC + SPP \cite{chen2020adaptive}   & 2020 & ResNet50 &332                          & 0.273& 0.648& /\\
&AFDC + SPP \cite{chen2020adaptive}   & 2020 & ResNet50 & (224, 256, 288, 320)          &0.271&0.649&0.671\\
&MUSIQ-single \cite{ke2021musiq}    & 2021 & MUSIQ&full                              &0.247&0.719&0.731\\
&MUSIQ \cite{ke2021musiq}           & 2021 & MUSIQ& (full, 384, 224)                 &0.242&0.726&0.738\\
&TANet \cite{he2022rethinking}      & 2022 & ResNet18 + MV2 & 224                    &/&0.758&0.765\\
&Celona et al. \cite{celona2022composition}      & 2022 & EfficientNet-B4 & 224                    &/&0.732&0.733\\
&Hou et al. \cite{hou2022distilling}      & 2022 &  [2* ResNeXt101 + ResNetv2] $\rightarrow$ ResNeXt101 &  300                   &/& 0.770 &0.770\\
&\shortstack{TAVAR \cite{li2023theme}}      & 2023 &2 * ResNet-50  + Swin-Base  & 224                    &/&0.725&0.736\\
&\shortstack{VILA-R \cite{ke2023vila}}      & 2023 &  ViT-B/16 & 224                    &/&\textcolor{sota}{\textbf{0.774}}&0.774\\
\midrule

\multirow{3}{*}{\shortstack{Supervised}} & \multirow{3}{*}{\shortstack{NIMA*}} 
& \multirow{3}{*}{2018} & MV2        &224   &0.314 & 0.661  & 0.669\\
& &  & ViT-Tiny        &224   &0.303 & 0.659  & 0.669\\
& &                     & Swin-Base  &224  & 0.241 & 0.754   & 0.762\\

\midrule
\multirow{3}{*}{\shortstack{Supervised}} & \multirow{3}{*}{CFA (ours)} 
& \multirow{3}{*}{-} & MV2        &224             & 0.293 & 0.679  & 0.688\\
& &                  & ViT-Tiny   &224             & 0.277  &0.690    & 0.701\\

& &                  & Swin-Base  &224             & 0.226 & 0.767   & 0.776\\

\midrule
\multirow{3}{*}{\shortstack{Semi-Supervised}} & \multirow{3}{*}{\shortstack{CSKD (ours) \\ (CFA + SKD)}} 
& \multirow{3}{*}{-} & Swin-Base $\rightarrow$ MV2  &224 &0.272 & 0.706  & 0.717\\
& &                  & Swin-Base $\rightarrow$ ViT-Tiny  &224   &  0.255 & 0.723  &0.734 \\

& &                  & Swin-Base $\rightarrow$ Swin-Base  &224   & \textcolor{sota} {\textbf{0.224}}  & {\textbf{0.770}}   & \textcolor{sota}{\textbf{0.779}}\\

\bottomrule
\end{tabular}
}
\label{tab:overall}
\end{table*}

\begin{table}[t]
\caption{Results on the PARA Dataset.}
\label{tab:para}
\centering
\resizebox{0.85\linewidth}{0.23\linewidth}{
\begin{tabular}{c|c|ccc}
 \toprule
Methods&\#Backbone & SRCC & PLCC \\   
\midrule

\multirow{2}{*}{\shortstack{TAVAR \cite{li2023theme}}}  &   2 * ResNet-50  + Mobilenet-v3  
   & 0.861 & 0.902                   \\
 &   2 * ResNet-50  + Swin-Base   &0.911  &0.940 \\
\midrule
\multirow{3}{*}{\shortstack{NIMA*}} &MV2&0.878&0.917  \\
   &ViT-Tiny&0.871&0.912  \\
   &Swin-Base&0.918&0.945  \\
\midrule
\multirow{3}{*}{\shortstack{CLIP}} &ViT-Large (lr=5e-6) &0.884 &0.918  \\
 &ViT-Large (lr=1e-5) &0.919 &0.947 \\
&ViT-Large (lr=5e-5) &0.851 &0.879 \\
\midrule
\multirow{3}{*}{\shortstack{CSKD (ours)}}  & Swin-Base $\rightarrow$ MV2   &  0.908 &  0.937 \\
     & Swin-Base $\rightarrow$ ViT-Tiny   &  0.913 &  0.942 \\
     & Swin-Base $\rightarrow$ Swin-Base   & \textcolor{sota}{\textbf{0.926}} &  \textcolor{sota}{\textbf{0.951}}       \\
 
\bottomrule
\end{tabular}
}
\end{table}
{\noindent\bf Evaluation Metrics}
Following previous work, mean squared error (MSE), Spearman rank order cross-correlation coefficient (SRCC) and Pearson linear cross-correlation coefficient (PLCC) are adopted as the performance indicators. MSE reflects the numeric distance between predictions and ground truth. PLCC and SRCC estimate the linear and ranking consistency between model estimations and human annotations. Both PLCC and SRCC range from -1 to 1, and higher values indicate better performance.

\subsection{Main Results}

{\noindent\bf Results on AVA Dataset}
Table \ref{tab:overall} presents a comprehensive comparison between the SOTA methods on the AVA dataset, and we include the size of the input image in the table for a fair comparison. We adopt different types of distillation in SKD: Self-distillation transfers knowledge from Swin-Base to Swin-Base, and big-to-small distillation transfers knowledge from a transformer to a lightweight CNN model (Swin-Base $\rightarrow$ MV2) or from a transformer to a lightweight transformer (Swin-Base $\rightarrow$ ViT-Tiny).

First, our models with different types of distillation demonstrate a substantial superiority over the SOTA methods under the comparison, confirming the effectiveness of the proposed CFA and SKD strategies. Second, the CFA model always yields significant improvements (up to 3.2\% in terms of PLCC) compared to NIMA*, the only difference between them lies in the pre-trained weights: in our CFA backbones are initialized with ImageNet pre-trained weights and then distilled based on CLIP, while in NIMA* backbones are initialized with ImageNet pre-trained weights only. This confirms our hypothesis that the aesthetic feature representation can benefit from the knowledge distillation based on CLIP. Third, comparing the CSKD models with CFA models validates the advantage of the use of semi-supervised KD to further enhance the performance of CFA backbones. Notably, Swin-Base $\rightarrow$ ViT-Tiny outperforms CFA model based on ViT-Tiny by 3.3\% in terms of SRCC and PLCC, validating the effectiveness of SKD in Section ~\ref{sec:skd}.

{\noindent\bf Results on AADB dataset}
The results of the proposed CSKD compared with three SOTA methods and NIMA* on the AADB dataset are summarized in Table A in Supplementary Material. We can see that our method achieves the best performance on all metrics, and the performance improvement compared to the SOTA methods is up to 6.8\% in terms of SRCC (0.726 vs 0.794).

{\noindent\bf Results on PARA dataset}
Table \ref{tab:para} shows the results on a recently published IAA benchmark, PARA. Our approach exhibits approximately 0.1 to 0.4 improvements in terms of SRCC/PLCC compared with the SOTA method TAVAR \cite{li2023theme} and the fine-tuned CLIP. Notably, directly fine-tuning CLIP is sensitive to the hyperparameter. 

\subsection{Interval Error Rate}
The distributions of most IAA datasets are Gaussian-like imbalanced or long-tailed, leading to local unreliability of global indicators such as SRCC. Hence, we propose an interval-based evaluation metric to precisely identify the local performance improvement. Specifically, we split the test set into different score intervals and calculate the interval error rate ($IER$) separately, 
\begin{equation}
    IER_k= \frac{1}{N_k}\sum_{i=1}^{N_k}\mathbb I(|(s^{pred}_i-s^{true}_i| > t), k = 0, 1,..., K-1.
\end{equation}
where $k$ and $K$ are the index of an interval and the total number of intervals, respectively, $N_k$ is the total number of samples in the interval $k$, $s_i^{pred}$ and $s_i^{true}$ are the predicted and real scores of the sample $i$, and $t=0.5$ is a fixed threshold, referring to the error tolerance. 

\begin{figure}[thb]
\centering
\includegraphics[width=0.65\linewidth]{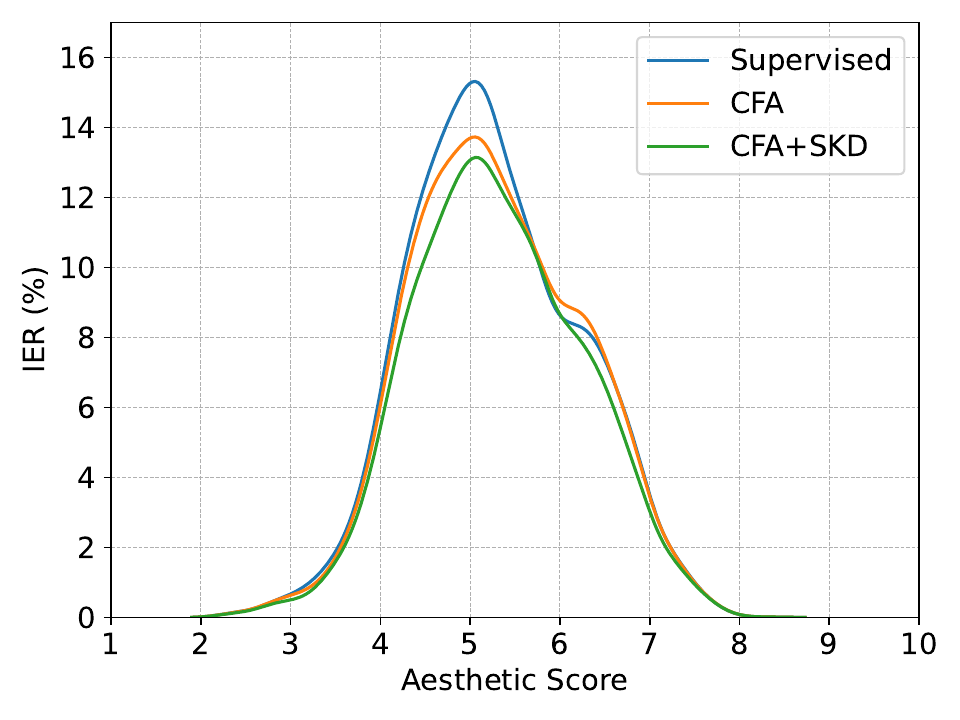}
\caption{Interval error rate visualization on AVA dataset.}
\label{fig:error}
\end{figure}

The $IER$ results on the AVA dataset are shown in Fig. \ref{fig:error}. First, the model derived from the CFA yields superior accuracy for images with medium IAA scores compared to the supervised baseline, confirming the efficacy of leveraging CLIP for feature representation in IAA. Second, employing the SKD to leverage the unlabeled data can further enhance the performance of well-trained CFA backbones. Finally, the proposed method achieves the overall best performance in terms of the $IER$.

\subsection{Ablation Study}

Ablation studies are conducted to verify the effectiveness of three key factors: 1) CLIP-based feature alignment, 2) multi-source unlabeled dataset, and 3) semi-supervised KD. Detailed analyses are provided in the Supplementary Material.

\section{Conclusion}

In this work, we propose CSKD, a unified two-phase CLIP-based semi-supervised knowledge distillation scheme for data-insufficient image aesthetics assessment tasks. Our method first distills rich features from CLIP to a lightweight IAA model with the help of a multi-source unlabeled dataset, aiming to enhance the representation ability of the widely-used visual backbones. Subsequently, to alleviate the reliance on large labeled datasets, we utilize semi-supervised knowledge distillation to leverage the unlabeled data and further enhance the knowledge transfer on the task-specific model. Extensive experiments on three IAA datasets demonstrated the superiority of the proposed method. We believe that the proposed CSKD shed light on the use of CLIP and knowledge distillation in IAA, and can be generalized to a variety of vision downstream tasks.

\section{Acknowledgement}
This work is sponsored by Shanghai Pujiang Program (23PJ1421800).

\bibliography{reference}

\begin{thebibliography}{10}

\bibitem{talebi2018learned}
Talebi et~al.,
\newblock ``Learned perceptual image enhancement,''
\newblock in {\em 2018 IEEE international conference on computational photography}. IEEE, 2018, pp. 1--13.

\bibitem{mai2016composition}
Mai et~al.,
\newblock ``Composition-preserving deep photo aesthetics assessment,''
\newblock in {\em CVPR}, 2016, pp. 497--506.

\bibitem{hosu2019effective}
Hosu et~al.,
\newblock ``Effective aesthetics prediction with multi-level spatially pooled features,''
\newblock in {\em CVPR}, 2019, pp. 9375--9383.

\bibitem{ke2021musiq}
Ke~et~al.,
\newblock ``Musiq: Multi-scale image quality transformer,''
\newblock in {\em CVPR}, 2021, pp. 5148--5157.

\bibitem{ke2006design}
Ke~et~al.,
\newblock ``The design of high-level features for photo quality assessment,''
\newblock in {\em CVPR}, 2006, pp. 419--426.

\bibitem{9777255}
Yang et~al.,
\newblock ``Metamp: Metalearning-based multipatch image aesthetics assessment,''
\newblock {\em IEEE Trans. Cybern.}, pp. 1--13, 2022.

\bibitem{she2021hierarchical}
She et~al.,
\newblock ``Hierarchical layout-aware graph convolutional network for unified aesthetics assessment,''
\newblock in {\em CVPR}, 2021, pp. 8475--8484.

\bibitem{hou2022distilling}
Hou et~al.,
\newblock ``Distilling knowledge from object classification to aesthetics assessment,''
\newblock {\em IEEE Trans. Circ. Syst. Video Tech.}, vol. 32, no. 11, pp. 7386--7402, 2022.

\bibitem{brown2020language}
Brown et~al.,
\newblock ``Language models are few-shot learners,''
\newblock {\em NeurIPS}, vol. 33, pp. 1877--1901, 2020.

\bibitem{yu2022coca}
Yu~et~al.,
\newblock ``Coca: Contrastive captioners are image-text foundation models,''
\newblock {\em arXiv preprint arXiv:2205.01917}, 2022.

\bibitem{radford2021learning}
Radford et~al.,
\newblock ``Learning transferable visual models from natural language supervision,''
\newblock in {\em ICML}. PMLR, 2021, pp. 8748--8763.

\bibitem{wang2023exploring}
Wang et~al.,
\newblock ``Exploring clip for assessing the look and feel of images,''
\newblock in {\em Proceedings of the AAAI Conference}, 2023, vol.~37, pp. 2555--2563.

\bibitem{ke2023vila}
Ke~et~al.,
\newblock ``{Vila}: Learning image aesthetics from user comments with vision-language pretraining,''
\newblock in {\em CVPR}, 2023, pp. 10041--10051.

\bibitem{zhang2023blind}
Zhang et~al.,
\newblock ``Blind image quality assessment via vision-language correspondence: A multitask learning perspective,''
\newblock in {\em CVPR}, 2023, pp. 14071--14081.

\bibitem{chen2020big}
Chen et~al.,
\newblock ``Big self-supervised models are strong semi-supervised learners,''
\newblock {\em NeurIPS}, vol. 33, pp. 22243--22255, 2020.

\bibitem{li2024graphadapter}
Li~et~al.,
\newblock ``Graphadapter: Tuning vision-language models with dual knowledge graph,''
\newblock {\em NeurIPS}, vol. 36, 2024.

\bibitem{wortsman2022robust}
Wortsman et~al,
\newblock ``Robust fine-tuning of zero-shot models,''
\newblock in {\em CVPR}, 2022, pp. 7959--7971.

\bibitem{yang2022personalized}
Yang et~al.,
\newblock ``Personalized image aesthetics assessment with rich attributes,''
\newblock in {\em CVPR}, 2022, pp. 19861--19869.

\bibitem{sandler2018mobilenetv2}
Sandler et~al.,
\newblock ``Mobilenetv2: Inverted residuals and linear bottlenecks,''
\newblock in {\em CVPR}, 2018, pp. 4510--4520.

\bibitem{deng2009imagenet}
Deng et~al.,
\newblock ``Imagenet: A large-scale hierarchical image database,''
\newblock in {\em CVPR}. Ieee, 2009, pp. 248--255.

\bibitem{zhou2022extract}
Zhou et~al.,
\newblock ``Extract free dense labels from clip,''
\newblock in {\em ECCV}. Springer, 2022, pp. 696--712.

\bibitem{rombach2022high}
Rombach et~al.,
\newblock ``High-resolution image synthesis with latent diffusion models,''
\newblock in {\em CVPR}, 2022, pp. 10684--10695.

\bibitem{fang2023eva}
Fang et~al.,
\newblock ``Eva: Exploring the limits of masked visual representation learning at scale,''
\newblock in {\em CVPR}, 2023, pp. 19358--19369.

\bibitem{hinton2015distilling}
Hinton et~al.,
\newblock ``Distilling the knowledge in a neural network,''
\newblock {\em arXiv preprint arXiv:1503.02531}, vol. 2, no. 7, 2015.

\bibitem{9340578}
L.~Wang and K.-J. Yoon,
\newblock ``Knowledge distillation and student-teacher learning for visual intelligence: A review and new outlooks,''
\newblock {\em IEEE Trans. Pattern Anal. Mach. Intell.}, vol. 44, no. 6, pp. 3048--3068, 2022.

\bibitem{wang2021knowledge}
Wang et~al.,
\newblock ``Knowledge distillation and student-teacher learning for visual intelligence: A review and new outlooks,''
\newblock {\em CVPR}, 2021.

\bibitem{yang2021survey}
X.~Yang, Z.~Song, I.~King, and Z.~Xu,
\newblock ``A survey on deep semi-supervised learning,''
\newblock {\em arXiv preprint arXiv:2103.00550}, 2021.

\bibitem{kuznetsova2020open}
Kuznetsova et~al.,
\newblock ``The open images dataset v4: Unified image classification, object detection, and visual relationship detection at scale,''
\newblock {\em International Journal of Computer Vision}, vol. 128, no. 7, pp. 1956--1981, 2020.

\bibitem{rw2019timm}
R.~Wightman,
\newblock ``Pytorch image models,'' \url{https://github.com/rwightman/pytorch-image-models}, 2019.

\bibitem{murray2012ava}
Murray et~al.,
\newblock ``Ava: A large-scale database for aesthetic visual analysis,''
\newblock in {\em CVPR}. IEEE, 2012, pp. 2408--2415.

\bibitem{kong2016photo}
Kong et~al.,
\newblock ``Photo aesthetics ranking network with attributes and content adaptation,''
\newblock in {\em ECCV}. Springer, 2016, pp. 662--679.

\bibitem{talebi2018nima}
H.~Talebi and P.~Milanfar,
\newblock ``Nima: Neural image assessment,''
\newblock {\em IEEE Trans. Image Process.}, vol. 27, no. 8, pp. 3998--4011, 2018.

\bibitem{touvron2021training}
Touvron et~al.,
\newblock ``Training data-efficient image transformers \& distillation through attention,''
\newblock in {\em ICML}. PMLR, 2021, pp. 10347--10357.

\bibitem{murray2017deep}
N.~Murray and A.~Gordo,
\newblock ``A deep architecture for unified aesthetic prediction,''
\newblock {\em arXiv preprint arXiv:1708.04890}, 2017.

\bibitem{sheng2018attention}
Sheng et~al.,
\newblock ``Attention-based multi-patch aggregation for image aesthetic assessment,''
\newblock in {\em ACM Multimedia}, 2018, pp. 879--886.

\bibitem{zeng2019unified}
Zeng et~al.,
\newblock ``A unified probabilistic formulation of image aesthetic assessment,''
\newblock {\em IEEE Trans. Image Process}, vol. 29, pp. 1548--1561, 2019.

\bibitem{chen2020adaptive}
Chen et~al.,
\newblock ``Adaptive fractional dilated convolution network for image aesthetics assessment,''
\newblock in {\em CVPR}, 2020, pp. 14114--14123.

\bibitem{he2022rethinking}
He~et~al.,
\newblock ``Rethinking image aesthetics assessment: Models, datasets and benchmarks,''
\newblock in {\em IJCAI}, 2022.

\bibitem{celona2022composition}
Celona et~al.,
\newblock ``Composition and style attributes guided image aesthetic assessment,''
\newblock {\em IEEE Trans. Image Process}, vol. 31, pp. 5009--5024, 2022.

\bibitem{li2023theme}
Li~et~al.,
\newblock ``Theme-aware visual attribute reasoning for image aesthetics assessment,''
\newblock {\em IEEE Trans. Circ. Syst. Video Tech.}, 2023.

\end{thebibliography}

\end{document}